\documentclass[11pt]{article}

\usepackage{authblk}
\usepackage{coling2020}
\usepackage{times}
\usepackage{url}
\usepackage{latexsym}
\usepackage{graphicx}
\usepackage{amsmath}
\usepackage{booktabs}
\usepackage{multirow}
\usepackage{subcaption}
\colingfinalcopy 

\title{Improvements and Extensions on Metaphor Detection}

\author[1]{Weicheng Ma}
\author[2]{Ruibo Liu}
\author[3]{Lili Wang}
\author[4]{Soroush Vosoughi}
\affil[ ]{Minds, Machines, and Society Group}
\affil[ ] {Department of Computer Science, Dartmouth College}
\affil[1,2,3]{\texttt{\{first.last.gr\}@dartmouth.edu}}
\affil[4]{\texttt{soroush.vosoughi@dartmouth.edu}}
\date{}

\begin{document}
\maketitle
\begin{abstract}
  Metaphors are ubiquitous in human language.
  The metaphor detection task (MD) aims at detecting and interpreting metaphors from written language,
  which is crucial in natural language understanding (NLU) research.
  In this paper, we introduce a pre-trained Transformer-based model into MD.
  Our model outperforms the previous state-of-the-art models by large margins in our evaluations,
  with relative improvements on the F-1 score from 5.33\% to 28.39\%.
  Second, we extend MD to a classification task about the metaphoricity of an entire piece of text to make MD applicable in more general NLU scenes.
  Finally, we clean up the improper or outdated annotations in one of the MD benchmark datasets and re-benchmark it with our Transformer-based model.
  This approach could be applied to other existing MD datasets as well, since the metaphoricity annotations in these benchmark datasets may be outdated.
  Future research efforts are also necessary to build an up-to-date and well-annotated dataset consisting of longer and more complex texts.
\end{abstract}

\section{Introduction}
\textit{Today we are drowning in a sea of social media posts.}
Metaphors serve as strong modifiers to the intentions and meanings of written texts.
In the header sentence, 
the metaphorical use of the word ``drown" in the sentence well expresses the worries of the speaker towards the large number of messages in social media, 
compared to the narrative version of the sentence, e.g. ``There are a lot of messages on social media".
As defined by Lakoff and Johnson~\shortcite{metaphor-definition}, metaphors involve words used outside their familiar domains.
For example, the word ``sea" in the leading sentence literally means a large body of water, 
but it is used metaphorically as a modifier to the phrase ``social media posts" to emphasize the abundance of messages in social media.
Similarly, people can ``drown" in water, but not in messages.
As shown in this example, metaphors are expressed by the context but not the aspect words themselves,
and there are no limits to the number of the metaphorical parts of speech.

Metaphor detection (MD) serves as a strong component in the natural language understanding (NLU) pipeline,
since NLU models cannot correctly process the meaning of written text without understanding the metaphors in the content.
MD serves to aid the NLU models by figuring out the metaphorical parts of speech in each sentence. 
However, this is a difficult task since metaphors are carried out by long spans of text, not by the appearance of single words or phrases. 
Existing algorithms and neural models are not able to encode long contexts without losing critical information related to metaphors. 
Moreover, the lack of labeled data and the difficulties in labeling metaphorical texts are obstacles to MD research as well. 
Due to these issues, the research on MD is still in an early stage and has not seen the improvements observed in other NLP tasks in recent years.

To reduce the annotation difficulties, 
researchers have been simplifying MD to a classification problem on the metaphoricity of one word or a word pair in each sentence.
Existing MD benchmark datasets are almost all labeled in this manner.
While the VUA dataset~\cite{vua} extends MD into a sequential labeling problem, it still limits the metaphorical parts-of-speech to be one per sentence.
This setting alleviates the pressure of early MD models which are based on handcrafted features
~\cite{imaginability,svm-crf-1,abstractness-1,domain-1,verbnet-1,imaginability-2}.
Nonetheless, the limitation overly simplifies MD and makes existing MD models inapplicable in NLU pipelines.
Since Rei et al.~\shortcite{similarity-net} first introduced deep learning to MD, 
recent models based on deep neural networks are already approaching the performance ceilings for the simplified version of MD.
Given the growing power of deep neural networks, it is time to re-define the task beyond the simplistic settings.

To verify our hypothesis, we fine-tune and evaluate a pre-trained BERT~\cite{bert} model on all the MD benchmark datasets.
Our model outperforms the previous state-of-the-art models with large margins, as expected.
The evaluation results almost all exceed 90\% in F-1 scores, 
suggesting that the existing MD settings and datasets are too easy for deep Transformer networks to solve.
We also extend MD to a classification task on the sentence level by removing the labels about the candidate metaphorical words.
While the results slightly drop on two MD datasets (0.32\% and 3.44\% in F-1 scores), they are still high, especially on trivial sentences.
We believe it is time to expand MD to include sentence-level metaphoricity labeling and to be evaluated on longer, more complex texts.

\begin{table}[t]
\centering
\begin{tabular}{|l|l|}
\hline
Sentence    & Her husband often \textbf{abuses} alcohol. \\ \hline
Explanation & To use excessively                         \\ \hline
Example     & Abuse alcohol                              \\ \hline
\end{tabular}
\caption{One example sentence from the MOH dataset that is wrongly labeled as metaphorical. The explanation of the word in bold and the example come from the Merriam-Webster dictionary.}
\label{tbl:wrong-annotation}
\end{table}

\begin{table*}[t]
\centering
\begin{tabular}{|l|l|l|}
\hline
Dataset & Sentence                                                                        & Label        \\ \hline
MOH     & He \textbf{absorbed} the knowledge or beliefs of his tribe.                              & Metaphorical \\ \hline
TroFi   & To expect banks to \textbf{absorb} a cost without a commensurate charge & \\
 & defies logic ./. & Non-Literal  \\ \hline
LCC     & Thank Lyndon Johnson, his Great Society, and the \textbf{War} on \textit{Poverty}.                & 3            \\ \hline
\end{tabular}
\caption{One example record in each of the three MD benchmark datasets. The bold words are the aspect words. In the LCC dataset, the target word (in italic) of the aspect word is also provided. The label sets are \{Literal, Metaphorical\} in the MOH dataset, \{Literal, Non-Literal\} in TroFi and \{0, 1, 2, 3\} in LCC.}
\label{tbl:dataset-example}
\end{table*}

In the evaluations, we uncover flaws in the MD benchmark datasets by analyzing the prediction errors our model makes.
One example of the annotation errors is displayed in Table \ref{tbl:wrong-annotation}.
While the word ``abuse" in this context literally means ``to use excessively", it is annotated as metaphorical in the MOH dataset.
The problematic annotations might result from recent updates to the dictionaries or changes in people's habits in using English.
This situation makes it difficult to label the benchmark datasets on the sentence level with the existing word-level annotations.
To validate our concerns about the quality of the annotations,
we clean up one of the MD benchmark datasets and have the new annotations checked by two native English speakers.
We also benchmark the re-annotated dataset with our model. 
The same strategy can and should be applied to other MD datasets to keep the annotations up to date.
We provide more details regarding the data analysis and re-annotation process in Section \ref{sct:discussion}.

The contributions of this paper are three-fold.
First, we report new state-of-the-art performances on three MD benchmark datasets to display the power of pre-trained deep Transformer networks on MD.
Second, we identify and clean up the annotation errors in one of the MD benchmark datasets through manual analysis and validation,
which will be made publicly available.
Third, we believe that the current settings of MD are overly simplistic for deep neural network models to solve, based on the evaluation performances of our model.
Thus, we extend MD to a sentence-level classification task and provide benchmark results on the three MD datasets.
Our future research efforts will involve the construction of an MD dataset with sentence-level annotations and longer and more complex texts.

\section{The Metaphor Detection Task}
Following Ge et al.~\shortcite{bilstm-attention}, we apply both the sequential labeling and word-level classification settings of MD in the experiments.
Also, we generalize the classification setting of MD to the sentence level, disregarding the aspect labels.
We describe the three settings of MD as follows.
For clarity, we use $s=\{w_1, w_2, ..., w_k\}$ to denote a sentence with $k$ words.\\
\textbf{Sequential labeling:} Given a sentence $s$, 
predict one label $l_i$ for each word $w_i$ indicating whether $w_i$ is metaphorical in the context.\\
\textbf{Word-level classification:} Given a sentence $s$ and an aspect word $w_i \in s$ (usually verbs, with exceptions),
predict the metaphoricity label $l_i$ associated with the aspect word.\\
\textbf{Sentence-level classification:} Given a sentence $s$,
predict whether $s$ is metaphorical.

The first two settings of MD have been extensively studied in previous research.
Since metaphors are expressed by the linguistic expressions,
attributing the metaphoricity of a sentence to an aspect word overly simplifies MD.
The sequential labeling setting makes it possible to evaluate the respective metaphoricity of each word in a sentence,
but annotating an MD dataset with complex sentences under the sequential labeling setting is too difficult and costly.
We provide the sentence-level classification formulation of MD for higher annotation quality while avoiding annotating an MD dataset on the token level.

\section{Datasets}
We base our evaluations and discussions on three MD benchmark datasets,
namely MOH~\cite{moh-dataset},
TroFi~\cite{trofi-dataset-1,trofi-dataset-2},
and LCC~\cite{lcc-dataset}.

The MOH dataset contains sentences from the WordNet~\cite{wordnet-1,wordnet-2} examples and the other two corpora are collected from news articles. 
The average number of words in the MOH dataset (7.40) is much lower than the TroFi (29.65) and LCC (28.66) datasets.
This makes the MOH dataset the simplest among the three benchmark datasets.
All three datasets provide one aspect word and a metaphoricity label for each sentence.
The label is associated with the aspect word.
The LCC dataset additionally provides the annotation about the target word of the aspect word in each sentence.
Different from the other two datasets, the LCC dataset annotates the metaphoricity scores of the aspect words in the set \{-1, 0, 1, 2, 3\}.
In the experiments, we get rid of the -1 labels in the LCC dataset since it denotes uncertain annotations.
We display one sample sentence from each dataset in Table \ref{tbl:dataset-example}.

The MOH dataset is constructed with 1640 sentences, 410 out of which are annotated as metaphorical.
The TroFi dataset is made of 1592 literal sentences and 2145 non-literal ones.
In the LCC dataset, 493 sentences are labeled as completely literal (0) while 1242, 1251 and 1838 sentences are annotated with metaphoricity scores of 1, 2, and 3, respectively.
We perform 10-fold cross-validation on all the three benchmark datasets under the word-level classification, sentence-level classification and sequential labeling settings in the experiments for fairness.

Though there exist other benchmark MD datasets as well, we choose to use the above three datasets intentionally.
The VUA dataset provides annotations for the sequential labeling setting of MD.
However, it is not publicly available now so we cannot obtain the data.
The TSV dataset~\cite{tsv} is also widely used, but its training set contains only a list of adjective-noun pairs without the context.
Despite the important role the aspect words play in MD, 
the lack of context content makes it improper to train or fine-tune deep Transformer-based models on the TSV dataset.
Clues for the sentence-level metaphoricity prediction cannot be learned in the training process either.
Thus we do not take these two datasets into our evaluation.

\section{Related Work}
Since MD is originally defined as a classification task, most early researchers solve it with logistic regression or SVM (Support Vector Machine) classifiers.
To use the information in the context, researchers concern much about the interrelations between the aspect words and the words closely related to them.
Thus POS (Part of Speech) tags and dependency paths are frequently used in the MD research.
Shutova et al.~\shortcite{unsupervised-1,unsupervised-2} cluster the grammatical relations between each pair of aspect word and its target word into clusters,
and they use rules to find out metaphorical combinations.
Topical information is also a crucial clue to the domain information of a sentence so it is widely used in MD.
Jang et al.~\shortcite{topical-1} represent the domain distribution of a sentence with sentence LDA.
They then base their metaphoricity predictions on the similarities, differences and transition patterns between adjacent sentence pairs.

It is interesting, though, that some words are regularly used metaphorically.
The intrinsic characteristics of these words are often taken into account when solving MD.
Strzalkowski et al.~\shortcite{imaginability} assume that highly imaginable words are promising metaphorical words.
They lookup the imaginability scores of the aspect words in the MRCPD lexicon~\cite{MRCPD-1,MRCPD-2} and label the words with high imaginability scores as metaphorical.
Similarly, Bracewell et al.~\shortcite{imaginability-2} also consider imaginability in predicting the metaphoricity of words.
Tsvetkov et al.~\shortcite{abstractness-1} and Turney et al.~\shortcite{abstractness-2} instead use abstractness of the aspect words or the entire sentences as features in detecting metaphors.
Other word-based features include the WordNet features (e.g. synonyms and semantic categories) \cite{imaginability,abstractness-1}, 
the VerbNet features (e.g. thematic roles) \cite{verbnet-1},
the domains of the candidate-words' arguments \cite{domain-1}, and the named entity information \cite{abstractness-1}.
Jang et al.~\shortcite{topical-1} claim that metaphors reveal the emotional or cognitive features of the author, so they use the occurrence of words in the LIWC lexicon~\cite{liwc} to model the sentences in their research.

Some researchers do not agree with the word-level classification setting of MD.
Instead, Hovy et al.~\shortcite{svm-crf-1} claim that every word in a sentence can be metaphorical or literal, 
making it unrealistic to list all the possible aspect words.
They introduce the sequential labeling setting of MD and apply CRF (Conditional Random Field) to solve it.
Researchers are actively studying MD as a sequential labeling task, but a well-annotated dataset under this setting is difficult to obtain at least for now.

Not until the year of 2016 did natural language processing (NLP) researchers start to use neural networks in MD.
With the power of neural networks, more and more researchers begin to examine the use of longer-term context information in MD.
Do Dinh et al.~\shortcite{mlp-1} encode the aspect words with an MLP (Multilayer Perceptron) taking the vectorized word embeddings, POS features and positional features as inputs.
They predict the metaphoricity of each aspect words by feeding their encodings into a logistic regression classifier.
Bulat et al.~\shortcite{embedding-1} similarly use pre-trained word embeddings to represent the aspect words,
and they use SVD (Singular Value Decomposition) to gain sentence representation for classification.
Shutova et al.~\shortcite{similarity-1} assume that the metaphoricity of a two-word phrase can be modeled with the cosine similarity between the aspect word embedding and the phrase embedding.
To represent the aspect word and phrase, they slide a window of fixed size on the context and use the information of all the words appearing in the window to encode the central word or the entire phrase.
They also introduce visual embeddings of words into MD which, according to their experimental results, help improve the results of MD on two benchmark datasets.
Rei et al.~\shortcite{similarity-net} extend the idea of Shutova et al.~\shortcite{similarity-1} by calculating a gated cosine similarity score between the two words' embeddings in each phrase with neural networks.
The research by Ge et al.~\shortcite{bilstm-attention} consider the entire sentence as useful context information and use BiLSTM with the attention mechanism to extract the features from the sentence automatically.
Most recently, Dankers et al.~\shortcite{bert-1} combine BERT with BiLSTM to jointly solve MD and the Emotion Regression task.
Their model yields good results on MD, but it does not fully exploit the encoding ability of BERT.
To go one step further, we design a BERT-based model and evaluate it on three standard evaluation datasets on MD in this paper.
\begin{figure}[t]
\centering
\begin{subfigure}{.4\textwidth}
    \centering
    \includegraphics[width=0.9\linewidth]{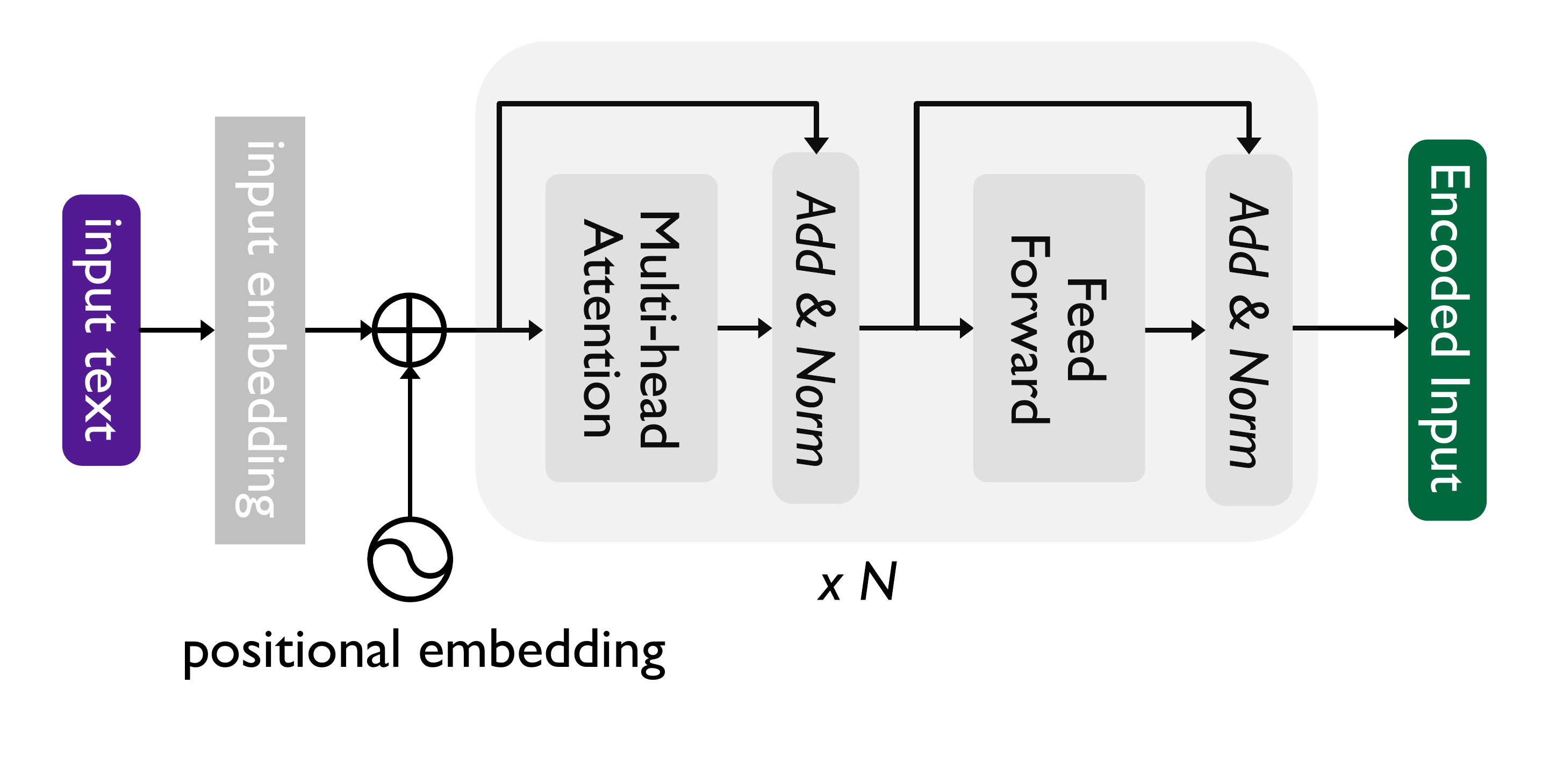}
    \caption{}
    \label{fig:transformer-architecture}
\end{subfigure}
\begin{subfigure}{.4\textwidth}
    \centering
    \includegraphics[width=0.9\linewidth]{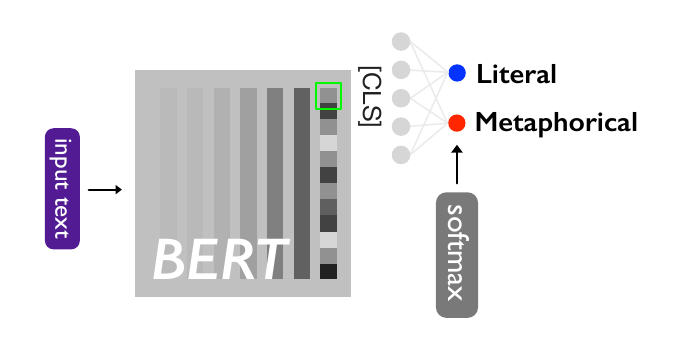}
    \caption{}
    \label{fig:model-architecture}
\end{subfigure}
\caption{The architecture of Transformer networks (a) and our model (b). N denotes the number of self-attention layers in a Transformer model.}
\end{figure}
\begin{table*}[t]
\centering
\begin{tabular}{|l|r|r|r|r|r|r|r|r|r|}
\hline
\multicolumn{1}{|c|}{\multirow{2}{*}{Model}} & \multicolumn{3}{c|}{MOH}                                                        & \multicolumn{3}{c|}{TroFi}                                                      & \multicolumn{3}{c|}{LCC}                                                        \\ \cline{2-10} 
\multicolumn{1}{|c|}{}                       & \multicolumn{1}{c|}{WCLS} & \multicolumn{1}{c|}{SCLS} & \multicolumn{1}{c|}{SL} & \multicolumn{1}{c|}{WCLS} & \multicolumn{1}{c|}{SCLS} & \multicolumn{1}{c|}{SL} & \multicolumn{1}{c|}{WCLS} & \multicolumn{1}{c|}{SCLS} & \multicolumn{1}{c|}{SL} \\ \hline
Dankers et al.~\shortcite{bert-1}                 & -                         & -                         & -                       & -                         & -                         & -                       & 76.90                     & -                         & -                       \\ \hline
Gao et al.~\shortcite{bilstm-attention}           & 79.10                     & -                         & 75.60                   & 72.00                     & -                         & 71.10                   & -                         & -                         & -                       \\ \hline
Shutova et al.~\shortcite{similarity-1}           & 75.00                     & -                         & -                       & -                         & -                         & -                       & -                         & -                         & -                       \\ \hline
Rei et al.~\shortcite{similarity-net}             & 74.20                     & -                         & -                       & -                         & -                         & -                       & -                         & -                         & -                       \\ \hline
BERT                                         & \textbf{85.52}            & \textbf{86.32}            & \textbf{89.18}          & \textbf{92.44}            & \textbf{92.12}            & \textbf{94.45}          & \textbf{81.00}            & \textbf{77.56}            & \textbf{91.48}          \\ \hline
\end{tabular}
\caption{Experimental results on the MOH, TroFi and LCC datasets with the word-level classification (WCLS), sentence-level classification (SCLS) and sequential labeling (SL) settings. All results are in terms of F-1 scores. BERT refers to our model.}
\label{tbl:experimental-results}
\end{table*}
\begin{figure}[h]
\begin{subfigure}{1.0\textwidth}
    \centering
    \includegraphics[width=0.9\linewidth]{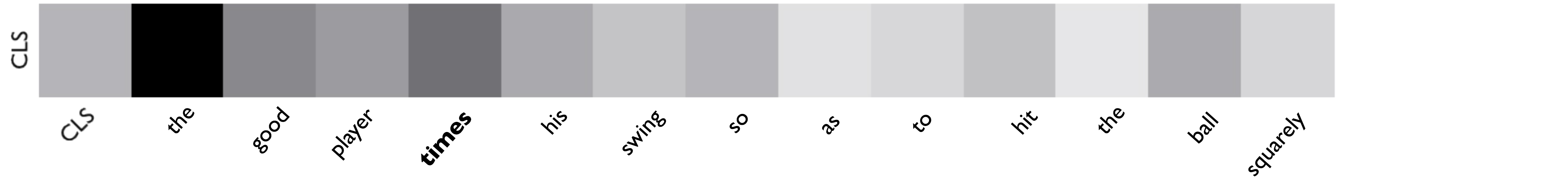}
    \caption{}
    \label{fig:moh-1}
\end{subfigure}
\begin{subfigure}{1.0\textwidth}
    \centering
    \includegraphics[width=0.9\linewidth]{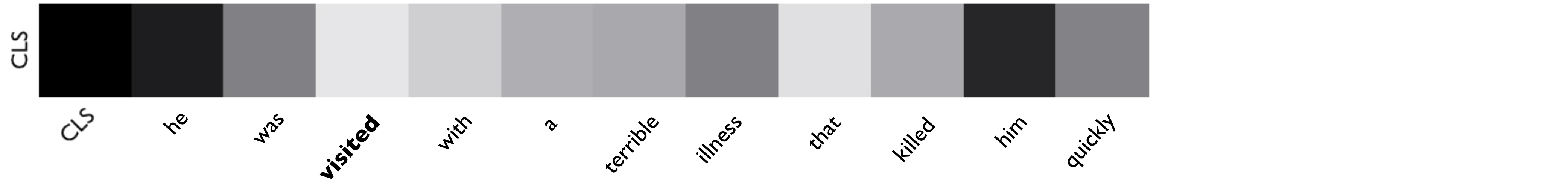}
    \caption{}
    \label{fig:moh-0}
\end{subfigure}
\begin{subfigure}{1.0\textwidth}
    \centering
    \includegraphics[width=0.9\linewidth]{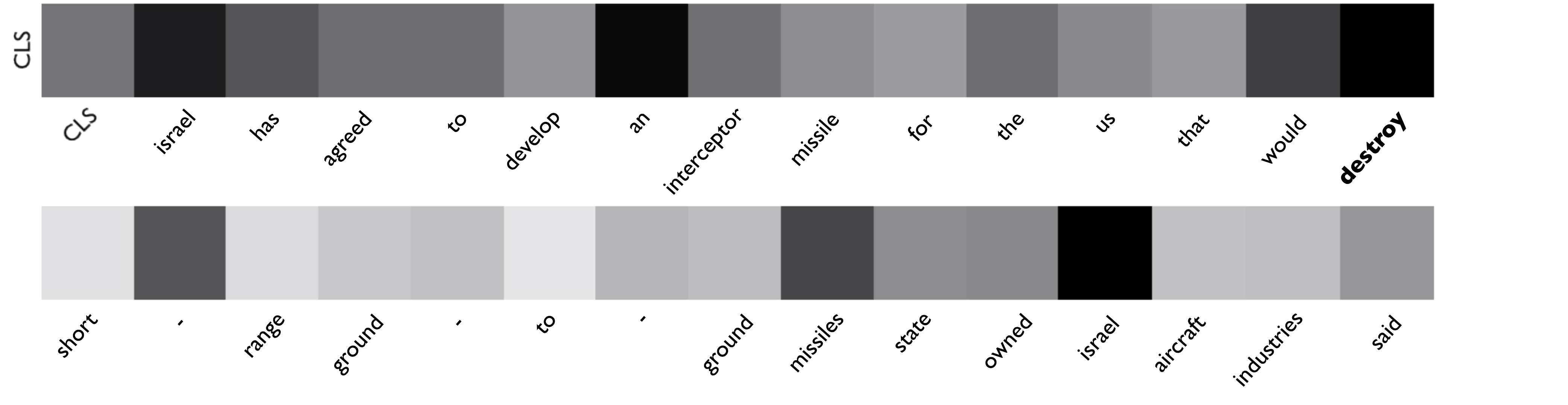}
    \caption{}
    \label{fig:trofi-1}
\end{subfigure}
\begin{subfigure}{1.0\textwidth} 
    \centering
    \includegraphics[width=0.9\linewidth]{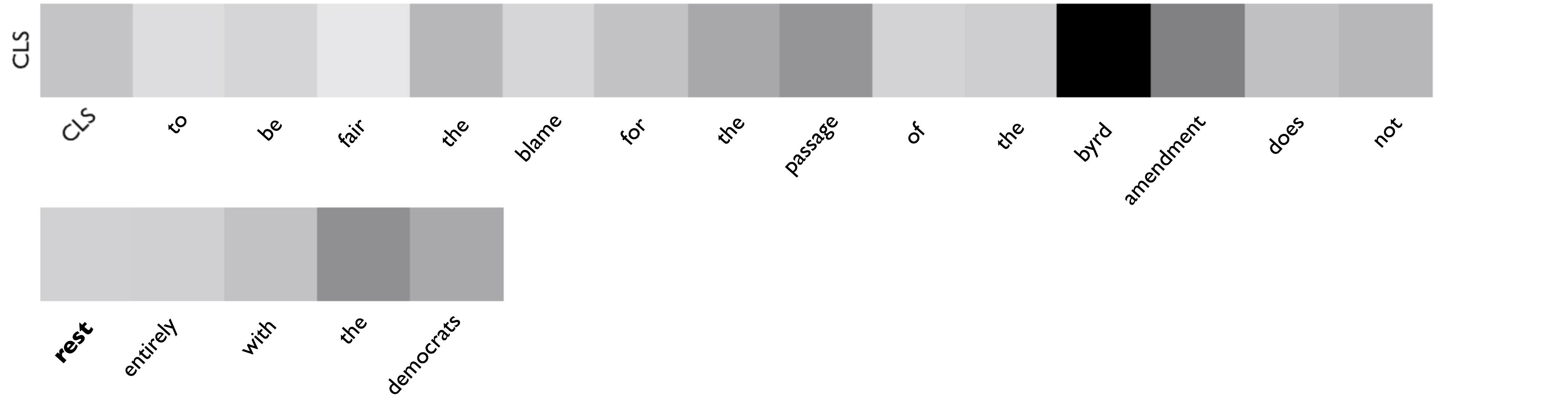}
    \caption{}
    \label{fig:trofi-0}
\end{subfigure}
\caption{Attention heatmaps generated by our sentence-level classification model on the MOH and TroFi datasets. The words in bold are the aspect words.}
\label{fig:attention-maps}
\end{figure}
\begin{figure}[h!]
\begin{subfigure}{1.0\textwidth}
    \centering
    \includegraphics[width=0.9\linewidth]{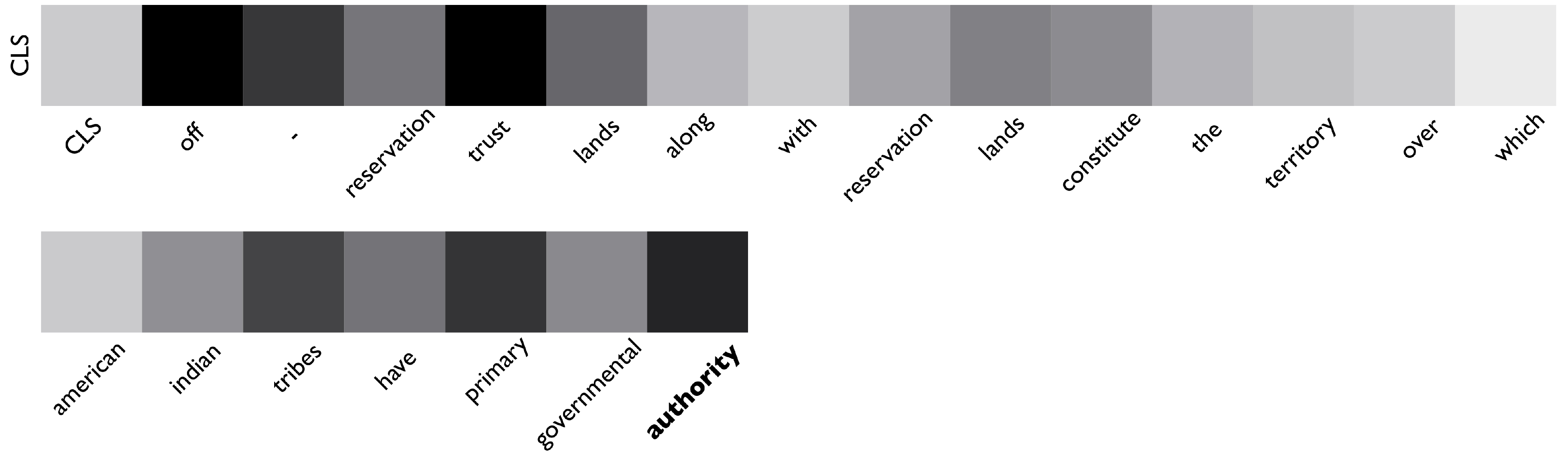}
    \caption{}
    \label{fig:lcc-0}
\end{subfigure}
\begin{subfigure}{1.0\textwidth}
    \centering
    \includegraphics[width=0.9\linewidth]{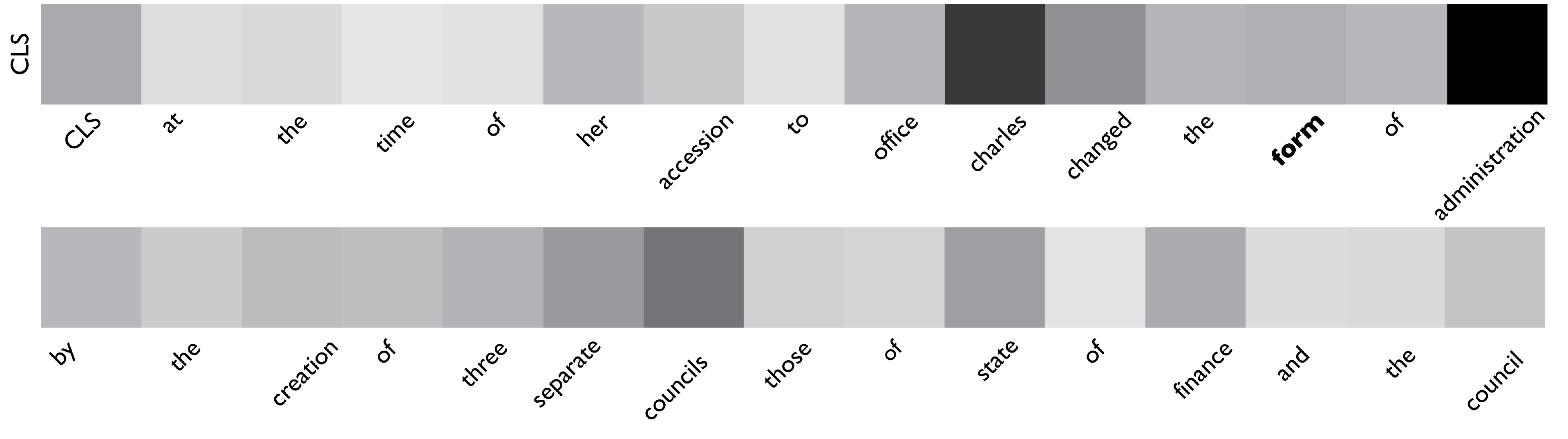}
    \caption{}
    \label{fig:lcc-1}
\end{subfigure}
\begin{subfigure}{1.0\textwidth}
    \centering
    \includegraphics[width=0.9\linewidth]{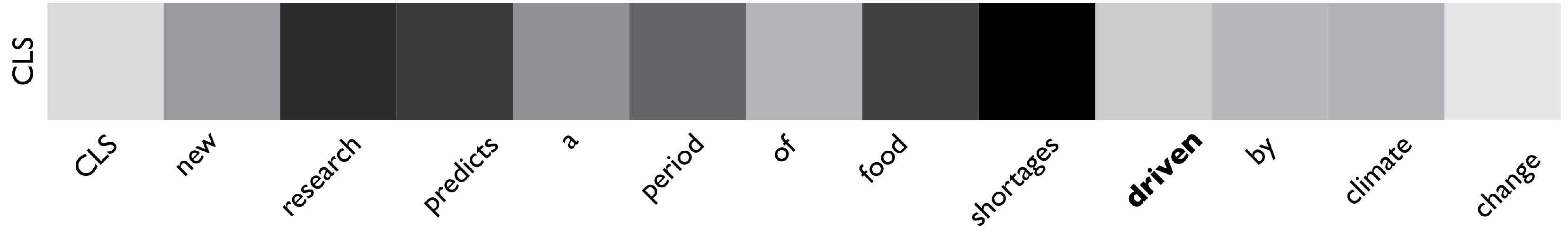}
    \caption{}
    \label{fig:lcc-2}
\end{subfigure}
\begin{subfigure}{1.0\textwidth}
    \centering
    \includegraphics[width=0.9\linewidth]{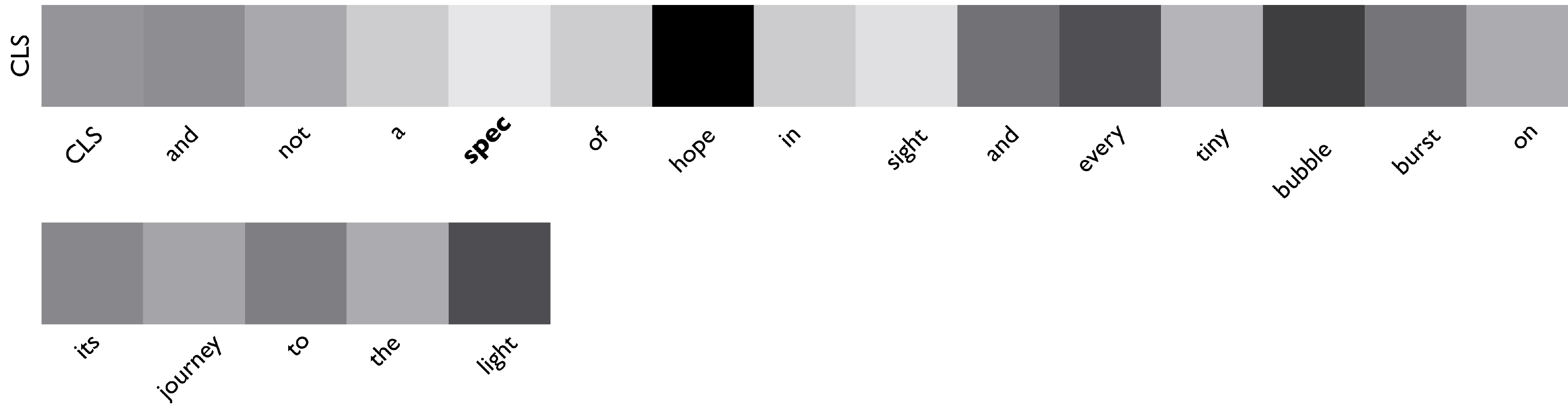}
    \caption{}
    \label{fig:lcc-3}
\end{subfigure}
\caption{Attention heatmaps generated by our sentence-level classification model on the LCC dataset.}
\label{fig:attention-maps-lcc}
\end{figure}

\section{Model Architecture}

The Transformer networks have been overtaking the state-of-the-arts in the NLP field since their emergence.
However, there have been very few works that have studied the usage of the Transformer networks in MD.
To the best of our knowledge, Dankers et al.~\shortcite{bert-1} made the first and only attempt in applying BERT~\cite{bert}, 
one of the most prevalent pre-trained Transformer-based models, on MD.
They build an MLP or additional attention layers on top of BERT to make metaphoricity predictions.
In our point of view, however, combining BERT with complex neural network architectures is a waste of its strength.
The additional layers co-trained with BERT are only exposed to the task-specific dataset which is much smaller than the BERT training data.
This makes it difficult to adapt BERT to the classification layers.
It is good enough to simply use a linear layer to resize the BERT output to the prediction space.
We specify the neural network architecture underlying BERT in Figure \ref{fig:transformer-architecture} and show in Figure \ref{fig:model-architecture} 
the simple model architecture with which we are able to achieve the state of the art on three MD benchmark datasets.
Our experiments are based on the PyTorch implementation of the Transformer networks by Huggingface~\cite{huggingface}.

\section{Experiments} \label{sct:experiments}
\begin{table*}[t]
\centering
\begin{tabular}{|l|l|l|l|l|}
\hline
Dataset                & ID              & Sentence                                                                    & Label                        & Pred               \\ \hline
\multirow{5}{*}{MOH}   & 1                  & The house \textbf{looks} north.                                             & metaphorical                 & literal                  \\ \cline{2-5} 
                       & 2                  & The huge waves \textbf{swallowed} the small boat and it sank &\multirow{2}{*}{metaphorical}                 & \multirow{2}{*}{literal} \\
                       & & shortly thereafter.     & &                  \\ \cline{2-5} 
                       & 3                  & You must \textbf{adhere} to the rules.                                               & metaphorical                 & literal                  \\ \cline{2-5} 
                       & 4                  & They \textbf{adhere} to their plan.                                                  & literal                      & literal                  \\ \hline
\multirow{6}{*}{TroFi} & 5                  & At 9 p.m . , a doctor \textbf{examines} her and orders tests ./.            & non-literal                  & literal                  \\ \cline{2-5} 
                       & \multirow{3}{*}{6} & The study , which \textbf{examined} 50 people who were  & & \\
                       & & wearing lap belts during auto accidents , concluded     & \multirow{3}{*}{non-literal} & \multirow{3}{*}{literal} \\
                       &                    &  that 32 would have `` fared substantially better &                              &                          \\
                       &                    & if they had been wearing a lap-shoulder belt . ''/''                        &                              &                          \\ \cline{2-5} 
                       & \multirow{2}{*}{7} & In order to focus federal resources on the SSC , its  &\multirow{2}{*}{non-literal} & \multirow{2}{*}{literal}\\
                       & & backers decided that Isabelle had to \textbf{die} ./.  &  &  \\\hline
\multirow{9}{*}{LCC}   & \multirow{6}{*}{8} & From this calculation it is obvious that with any \textbf{form} & & \\
                       & & of taxation per head the State is baling out the last & \multirow{6}{*}{2}           & \multirow{6}{*}{3}       \\
                       &                    &  coppers of the poor taxpayers in order to settle accounts &                              &                          \\
                       &                    &  with wealthy foreigners, from whom it has borrowed &                              &                          \\
                       &                    &  money instead of collecting these coppers for its own      &                              &                          \\
                       &                    & needs without the additional interest.                                                        &                              &                          \\ \cline{2-5} 
                       & \multirow{2}{*}{9} & The organism that causes gonorrhea (gonococcus) is an       & \multirow{2}{*}{2}           & \multirow{2}{*}{3}       \\
                       &                    & \textbf{example} of a bacterial invader.                                                     &                              &                          \\ \cline{2-5} 
                       & \multirow{2}{*}{10}                 & Background Checks - Local Background Checks Can     & \multirow{2}{*}{1}                            & \multirow{2}{*}{0}                        \\
                       & & \textbf{Reduce} Deaths. & & \\\hline
\end{tabular}
\caption{Example prediction errors on the MOH, TroFi and LCC datasets. The source words are in bold.}
\label{tbl:prediction-errors}
\end{table*}
As is mentioned in previous sections, 
we fine-tune and evaluate BERT models for classification and for sequential labeling on the Trofi, MOH and LCC datasets with 10-fold cross validation.
In the experiments, we use the pre-trained \textbf{bert-base-cased} model released by Google.
The model architecture is a 12-layer Transformer model with 12 attention heads on each layer.
The hidden dimension of the model is 768.
We limit the sentence lengths to 128 since it fits most of the sentences in the three datasets.
In both the fine-tuning and evaluation process, we set the batch size to 128.
As for training epochs, we use 5 for the aspect-based classification setting, 20 for the sentence-based setting and 20 for the sequential labeling formulation.
We select the training epochs through manually monitoring the training process to avoid overfitting.

Our evaluation is performed under the three MD settings respectively.
For the word-based classification setting, we mask out the aspect word in each sentence and concatenate the pair of sentences with and without the mask as input.
In this way, we take advantage of BERT's next sentence prediction mechanism.
Since BERT infers the masked words with contextual information, 
it is highly probable that the masked word is used literally if the two sentences are predicted to be in the same context.
In the sentence-level formulation, we directly feed into the model the original sentence without any change.
The sequential labeling model takes the words and their indexes in the sentence as input and predicts the metaphoricity label of each word.
We label the aspect words with their annotated labels and regard all the other words as literally used in the evaluation.

Table \ref{tbl:experimental-results} displays the 10-fold cross-validation results of our model and the baseline models on the three benchmark datasets.
Our model outperforms the baseline models by large margins and constructs the new state of the art under all the three settings.
The success of the models based on Elmo~\cite{elmo} and BERT demonstrates the importance of contextual information in MD.
By comparing our model to that of Ge et al.~\shortcite{bilstm-attention} which relies on Elmo embeddings, we demonstrate the outstanding encoding ability of BERT.
Though both based on the BERT model, our model shows superior performance in MD than that of Dankers et al.~\shortcite{bert-1}.
This supports our assumption that overly complex classifiers built on top of BERT negatively affect the fine-tuning process.

The results show that in most cases, our model performs the best in the word-based classification setting.
The more complex the sentences in the datasets are (LCC $>$ TroFi $>$ MOH), 
the more difficult the sentence-based classification setting of MD is than the word-based classification setting.
This agrees with our expectation since there can be multiple metaphorical words in a sentence that influences the prediction of our model.
Our model performs surprisingly well on the TroFi dataset, even better than on the MOH dataset.
This might be due to the difficulty of training deep neural models on the overly simple sentences in the MOH dataset.
Our model shows great potential under the sequential labeling setting as well.
On all the three datasets, our model achieves F-1 scores close to or even above 90\%.
We are highly impressed by the power of the BERT model and we feel that the existing MD benchmark datasets are becoming too easy for deep Transformer-based models to solve.
So it is time to construct new corpora containing longer and more complex text with multiple metaphorical components in each piece of text.
By extending the MD research to more complex realistic scenes, the MD models can better aid the NLU research and benefit the NLP community.

\section{Analysis and Discussions} \label{sct:discussion}
We manually inspect the predictions our model makes to analyze the causes of the prediction errors.
Table \ref{tbl:prediction-errors} displays typical prediction errors in our evaluation.
The major problem with the MOH dataset is the unbalanced labels for each aspect word.
Grouping by the aspect words, 194 out of the 438 word groups in the MOH dataset contain no metaphorical annotations and 11 groups have no literal annotations.
The labels in the rest word groups are not balanced either.
Models trained on unbalanced training data are likely to associate the label predictions with the appearance of the aspect words.
Sentence 1 in Table \ref{tbl:prediction-errors} is the only metaphorical record with the aspect word ``look" in the MOH dataset, for example.
The model might have learned to classify all the sentences with the verb ``look" into the literal class, generating this error case.
On the other hand, most sentences in the MOH dataset are simple and the aspect words are often the only verbs.
This increases the difficulty of our model to generalize the learned knowledge into predictions on longer and more complex sentences in the validation dataset.
Sentence 2 features the metaphorical word ``swallow" but our model is disturbed by the literal word ``sink" and makes the wrong prediction.
As all the sentences with ``swallow" in the MOH dataset are annotated as metaphorical, 
this prediction error proves that our model learns to classify not from the single aspect words, but a global view of the sentences.
Some annotations in the MOH dataset are difficult for us to understand.
For instance, ``adhere to the rules" in Sentence 3 is labeled as metaphorical while ``adhere to the plan" in Sentence 4 is literal.
This leads to our hypothesis that the annotations may be wrong or outdated.
With this idea in mind, we re-annotated the MOH dataset.
In the resulted dataset, 402 out of the 1639 annotations (24.53\%) are different from the original labels.
To alleviate the problem caused by the subjectivity in the metaphoricity annotations,
we sampled 100 from the records where our annotations do not agree with the original ones and had it validated with three native speakers.
The agreement rate of the three independent annotators on the new annotations is 66\%.
This proves that our annotations are better in quality than the original labels.
We use majority vote to re-label the MOH dataset and benchmark the revised dataset with our BERT-based model.
The 10-fold cross-validation results are 94.21\%, 94.21\%, and 98.22\% under the word-level classification, sentence-level classification and sequential labeling settings, respectively.

Our model performs much better on the TroFi dataset than on the MOH dataset, 
benefited from the abundant instances in each word group and the relatively balanced labels.
However, we do not fully agree with the annotations either.
The label for the word ``examine" in Sentence 5, for example, is metaphorical,
though the usage of ``examine" in this sentence well aligns with its literal meaning ``test or examine for the presence of disease or infection".
Similarly, the ``examine" in Sentence 6 is used in its literal meaning ``to question or examine thoroughly and closely",
but it is labeled as metaphorical.
Since the TroFi dataset is collected from news articles, abbreviations sometimes cause trouble in the evaluation as well.
The name ``Isabelle" in Sentence 7 can well denote a person without preliminary knowledge about SSC (Superconducting Supercollider) in the context.
It is then understandable why our model predicts the sentence as using the verb ``die" literally.
In the future, 
we suggest adding the surrounding sentences in the context into the dataset to make MD better defined and more appropriate for training deep neural network models.

Different from the MOH and TroFi datasets, the LCC dataset does not limit the source words to verbs.
Another difference is that the labels in the LCC dataset are metaphoricity scores.
This makes the LCC dataset more difficult to solve.
Our model predicts 3 while the label is 2 for the word ``form" in Sentence 8, for example.
Possibly our model detects the metaphorical use of ``copper" in the same sentence and decides to assign a higher metaphoricity score to the entire sentence.
The prediction error of our model in Sentence 9 is in a similar case.
Our model predicts a high score due to the synergy of the metaphorical words ``example" and ``invader".
The annotations in the LCC dataset are sometimes controversial as well.
The word ``reduce" in Sentence 10 perfectly matches the literal meaning ``to cut down on", but is annotated as 1 (weakly metaphorical) in LCC, for example.

On the other hand, since higher attention weights are put on the evidence for classification in Transformer-based models,
we examine the attention maps on the last self-attention layer generated by our model under the sentence-level classification setting to interpret the performance of our model.
Under the sentence-level classification setting, the predictions are made from the hidden states of the CLS token.
So we evaluate the attention scores of the CLS token on all the other words in each sentence.
The rule well applies to the word-level classification and sequential labeling settings, only with different tokens on which to base the predictions.
To avoid duplication, we only display the attention heatmaps generated under the sentence-level setting in this paper.
Figure \ref{fig:attention-maps} displays the attention heatmaps on MOH and TroFi examples to reflect the influence of metaphorical polarity on the attention scores and Figure \ref{fig:attention-maps-lcc} contains the heatmaps on LCC examples to show the effect of metaphorical intensities.
In Figure \ref{fig:moh-1}, the subject ``the good player", the verb ``times" and the object ``his swing" are all heavily attended, indicating the literal usage of the word ``time" paired with the words ``player" and ``swing".
Quite on the contrary, the verb ``visited" in Figure \ref{fig:moh-0} is very lightly attended compared to ``he" and ``illness" in the same sentence, which is a signal of the metaphorical use of the word ``visited" in its context.
The same pattern applies to the examples in TroFi (Figure \ref{fig:trofi-0} and \ref{fig:trofi-1}) and LCC (Figure \ref{fig:lcc-0}, \ref{fig:lcc-1}, \ref{fig:lcc-2} and \ref{fig:lcc-3}) that the more heavily our model attends on an aspect word, the lower chance it is used metaphorically in the context.
It is worth noting that when the sentences grow longer, the amount of potential aspect words also increases.
The use of these aspect words can be literal or metaphorical at the same time, 
which benefits classifying the metaphoricity of the sentence as a whole.
In Figure \ref{fig:moh-1}, for instance, the verb ``hit" is used literally with the noun ``ball" as well.
But there are also cases where the multiple aspect words in one sentence hold different metaphoricities, e.g. the words ``swallow" and ``sink" in Sentence 2 of Table \ref{tbl:prediction-errors}.
These examples contribute to many prediction errors made by our sentence-level classification model but are generally not a problem for the aspect-based classification and sequential labeling models.
As we stated before, examining the metaphoricity of given aspect words only simplifies MD.
Given the powerful neural models in the NLP field, we do not need this type of simplification anymore.
As our next step, we will keep working on labeling MD datasets on the sentence level or on the aspect level with multiple aspect words per sentence.
We will also introduce social media data to MD for richer metaphorical expressions and varied topics.


\section{Conclusion and Future Work}
Though difficult, MD has been an important task in the NLP community.
In this paper, we refined the definitions of MD by defining a new task formulation.
We also designed and evaluated a BERT-based model on three MD benchmark datasets.
Our model largely outperformed the previous state-of-the-art methods.
Through analysis of the prediction errors made by our model, 
we found that a large number of prediction errors can be attributed to the simplicity of the datasets and the annotation qualities.
To validate this, we re-annotated the MOH dataset and manually verified the quality of our new annotations.
We saw in the experiments that our model achieves very high accuracy on existing MD benchmark datasets, 
meaning that they are becoming overly simple for deep neural networks.
Our future work will focus on collecting and annotating a new MD dataset with more complex texts.
Regarding the prosperity of social media, we also plan to address the metaphor detection problem on informal text.
We hope our work will attract more research interest to MD and we call for future contributions to solve the problem.

\bibliographystyle{coling}
\bibliography{coling2020}

\end{document}